\documentclass[letterpaper, 10 pt, conference]{ieeeconf}
\IEEEoverridecommandlockouts    % This command is only needed if
\usepackage{graphics}           
\usepackage{times}              
\usepackage{amsmath}            
\usepackage{amssymb}            
\usepackage{graphicx}
\usepackage{algorithm}
\usepackage[noend]{algpseudocode}
\usepackage{booktabs}
\usepackage{color}
\definecolor{instructioncolor}{rgb}{.5,.5,.5}

\usepackage[font=small]{caption}

\def\secref#1{Sec.~\ref{#1}}
\def\figref#1{Fig.~\ref{#1}}
\def\tabref#1{Tab.~\ref{#1}}
\def\eqref#1{Eq.~(\ref{#1})}
\def\algref#1{Alg.~\ref{#1}}

\makeatletter
\usepackage{xspace}
\DeclareRobustCommand\onedot{\futurelet\@let@token\@onedot}
\def\@onedot{\ifx\@let@token.\else.\null\fi\xspace}
\def\eg{e.g\onedot}

\def\etc{etc\onedot} 
 
\def\etal{{et al}\onedot}
\makeatother

\usepackage{array}
\newcolumntype{L}[1]{>{\raggedright\let\newline\\\arraybackslash\hspace{0pt}}m{#1}}
\newcolumntype{C}[1]{>{\centering\let\newline\\\arraybackslash\hspace{0pt}}m{#1}}
\newcolumntype{R}[1]{>{\raggedleft\let\newline\\\arraybackslash\hspace{0pt}}m{#1}}

\newcommand{\RR}{\mathbb{R}}

\renewcommand{\v}[1]{{\b #1}} 

\usepackage{siunitx}
\usepackage{subcaption}
\usepackage{graphics}    % for pdf, bitmapped graphics files
\usepackage{times}       % assumes new font selection scheme installed
\usepackage{amsmath}     % assumes amsmath package installed
\usepackage{amssymb}     % assumes amsmath package installed
\usepackage{graphicx}
\usepackage{xcolor}
\usepackage[algo2e, linesnumbered,ruled,noline]{algorithm2e} 
\usepackage{multirow}
\usepackage[colorlinks,linkcolor=blue]{hyperref}

\DeclareMathOperator{\RMSE}{RMSE}

\renewcommand{\v}[1]{{\mathbf{#1}}}

\SetKwInput{KwInput}{Input}                % Set the Input
\SetKwInput{KwOutput}{Output} 
\title{\LARGE \bf Online Range Image-based Pole Extractor \\ for Long-term LiDAR Localization in Urban Environments}

\author{Hao Dong$^*$ \and Xieyuanli Chen$^*$ \and Cyrill Stachniss% <-this % stops a space
  \thanks{$^*$Authors contributed equally.
  H. Dong is with the Aalto University, Finland.
  X. Chen and C. Stachniss are with the University of Bonn, Germany. This work has partially been funded by the European Union’s Horizon 2020 research and innovation programme under grant agreement No~101017008~(Harmony), and by the Chinese Scholarship Committee.
  \newline 978-1-6654-1213-1/21/\$31.00 \textcopyright 2021 IEEE}%
}

\begin{document}
\maketitle
\thispagestyle{empty}
\pagestyle{empty}

%%%%%%%%%%%%%%%%%%%%%%%%%%%%%%%%%%%%%%%%%%%%%%%%%%%%%%%%%%%%%%%%%%%%%%%%%%%%%%%%
\begin{abstract}
  Reliable and accurate localization is crucial for mobile autonomous systems. 
  Pole-like objects, such as traffic signs, poles, lamps, \etc, are ideal landmarks for localization in urban environments due to their local distinctiveness and long-term stability. 
  In this paper, we present a novel, accurate, and fast pole extraction approach that runs online and has little computational demands such that this information can be used for a localization system. 
  Our method performs all computations directly on range images generated from 3D LiDAR scans, which avoids processing 3D point cloud explicitly and enables fast pole extraction for each scan. 
  We test the proposed pole extraction and localization approach on different datasets with different LiDAR scanners, weather conditions, routes, and seasonal changes. 
  The experimental results show that our approach outperforms other state-of-the-art approaches, while running online without a GPU. 
  Besides, we release our pole dataset to the public for evaluating the performance of pole extractor, as well as the implementation of our approach. 
\end{abstract}

%%%%%%%%%%%%%%%%%%%%%%%%%%%%%%%%%%%%%%%%%%%%%%%%%%%%%%%%%%%%%%%%%%%%%%%%%%%%%%%%
\section{Introduction}
\label{sec:intro}

Accurate and reliable localization is a basic requirement for an autonomous robot. 
The accurate estimation of its state helps the robot to avoid collision with the obstacles, follow the traffic lanes, and perform other tasks. 
Reliability means here that the robot should adapt to changes in the environment, such as different weather, day and night, seasonal changes, etc.

GPS or GNSS-based localization systems are robust to appearance changes of the environments. 
However, in urban areas, they may suffer from low availability due to building and tree occlusions. 
Additional, map-based approaches are needed for precise and reliable localization for mobile robots. 
Multiple different types of sensors have been used to build the map of the environments, \eg LiDAR scanners~\cite{droeschel2018icra, behley2018rss, vizzo2021icra}, monocular cameras~\cite{mur-artal2015tro}, stereo cameras~\cite{cvisic2017jfr}, \etc. 
Among them, LiDAR sensors are more robust to the illumination changes, and multiple LiDAR-based effective and efficient mapping approaches have been proposed, for example, the work by Droeschel and Behnke~\cite{droeschel2018icra} and by Behley and Stachniss~\cite{behley2018rss}.
However, these approaches often need large amounts of memory due to map representations, thus cannot generalize easily to large-scale scenes. 
If only specific features are used to build the map, such as traffic signs, trunks and other pole-like structures, the map size can be reduced significantly. 

\begin{figure}[t]
	\centering
	\begin{subfigure}[b]{\linewidth}
	  \includegraphics[width=\linewidth]{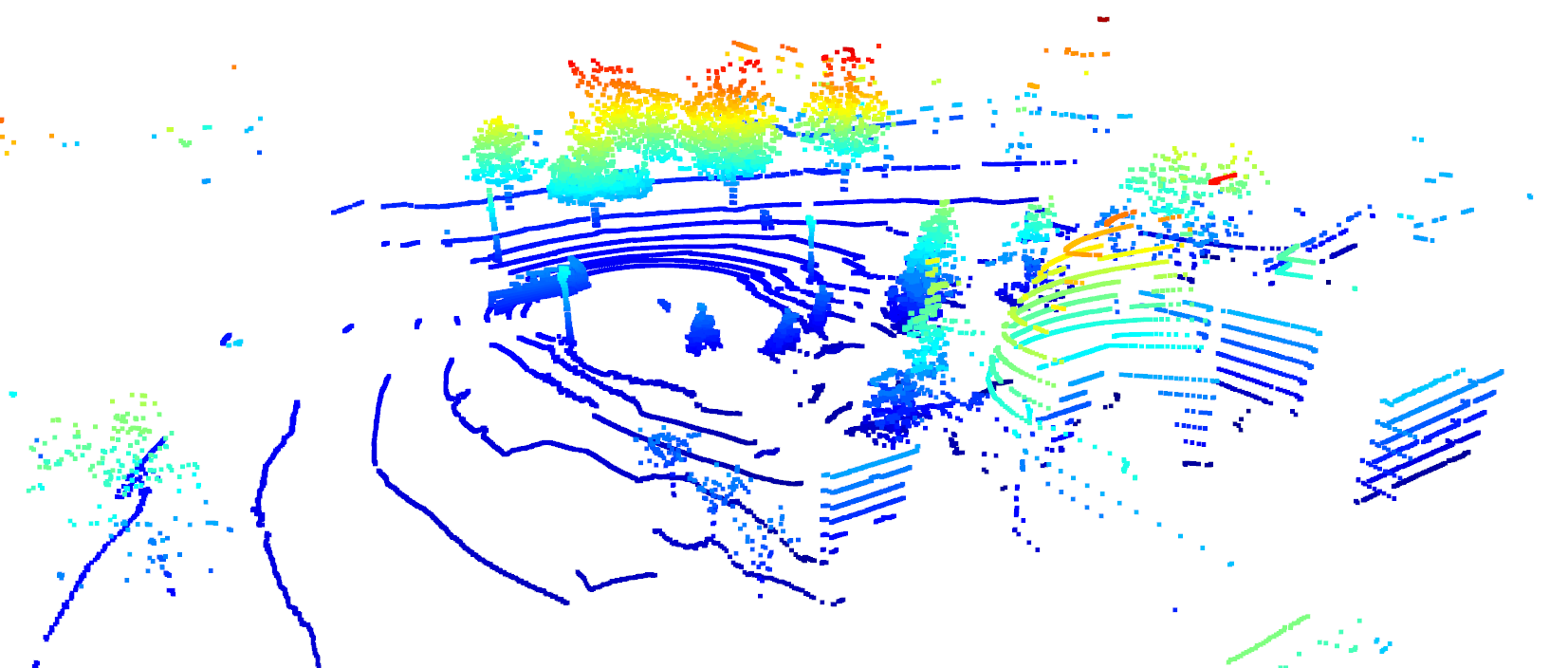}
	  \caption{Current scan}
	\end{subfigure}
	\begin{subfigure}[b]{\linewidth}
	  \includegraphics[width=\linewidth]{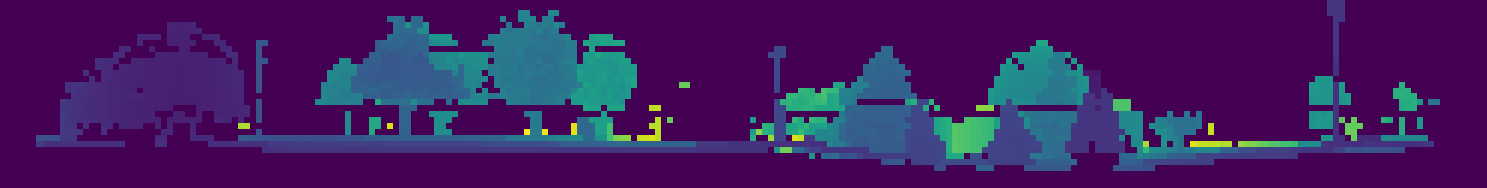}
	  \caption{Current range image}
	\end{subfigure}
	\begin{subfigure}[b]{\linewidth}
		\includegraphics[width=\linewidth]{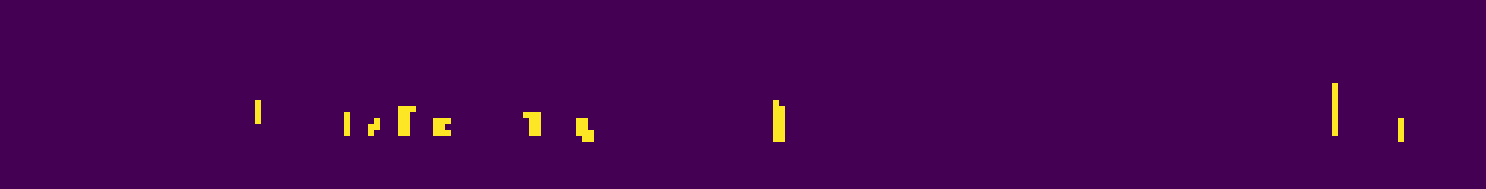}
		\caption{Pole extraction result by our approach}
	  \end{subfigure}
	\caption{Visualization of range image and pole extraction result. On the top is the raw LiDAR scan. The corresponding range image generated from this scan is in the middle. 
	The bottom is the pole extraction result based on range image.}
	\label{fig:pole_extractor}
  \end{figure}

The main contribution of this paper is a novel range image-based pole extractor for long-term localization of autonomous mobile systems. 
Instead of using directly the raw point clouds obtained from 3D LiDAR sensors, we investigate the use of range images for pole extraction. 
Range image is a light and natural representation of the scan from a rotating 3D LiDAR such as a Velodyne or Ouster sensors. 
Operating on such an image is considerably faster than on the raw 3D point cloud. 
Besides, range image keeps the neighborhood information implicitly in its 2D structure and we can use this information for segmentation.
As shown in \figref{fig:pole_extractor}, in the mapping phase, we first project the raw point cloud into a range image and then extract poles from that image. 
After obtaining the position of poles in the range image, we use the ground-truth poses of the robot to reproject them into the global coordinate system to build a global map. 
During localization, we utilize Monte Carlo localization~(MCL) for updating the importance weights of the particles by matching the poles detected from online sensor data with the poles in the global map. 

In sum, we make three key claims:
Our approach is able to
(i)~extract more reliable poles in the environment compared to the baseline method,
(ii)~as a result, achieve better localization performance in different environments, and 
(iii)~run online without a GPU.
These claims are backed up by the paper and our experimental evaluation.
The source code of our approach together with all parameters and the pole dataset can be accessed at:
\href{https://github.com/PRBonn/pole-localization}{https://github.com/PRBonn/pole-localization}.

%%%%%%%%%%%%%%%%%%%%%%%%%%%%%%%%%%%%%%%%%%%%%%%%%%%%%%%%%%%%%%%%%%%%%%%%%%%%%%%%
\section{Related Work}
\label{sec:related}

For localization given a map, there exists a large amount of research.
While a lot of different types of sensors have been used to tackle this problem~\cite{thrun2005probrobbook}, in this work, we mainly concentrate on LiDAR-based approaches.

Traditional approaches to robot localization rely on probabilistic state estimation techniques. 
A popular framework is Monte Carlo localization~\cite{dellaert1999icra}, which uses a particle filter to estimate the pose of the robot and is still widely used in robot localization systems~\cite{chen2020iros,chen2021icra}.

Besides the traditional geometric-based methods, more and more approaches recently exploit deep neural networks and semantic information for 3D LiDAR localization. 
For example, Ma \etal~\cite{ma2019iros} combine semantic information such as lanes and traffic signs in a Bayesian filtering framework to achieve accurate and robust localization within sparse HD maps, whereas Tinchev \etal~\cite{tinchev2019ral} propose a learning-based method to match segments of trees and localize in both urban and natural environments.
Sun \etal~\cite{sun2020icra} use a deep-probabilistic model to accelerate the initialization of the Monte Carlo localization and achieve a fast localization in outdoor environments.
Wiesmann \etal~\cite{wiesmann2021ral} propose a deep learning-based 3D network to compress the LiDAR point cloud, which can be used for large-scale LiDAR localization.
In our previous work~\cite{chen2020iros, chen2020rss}, we also exploit CNNs with semantics to predict the overlap between LiDAR scans as well as their yaw angle offset, and use this information to build a learning-based observation model for Monte Carlo localization.
The learning-based methods perform well in the trained environments, while they usually cannot generalize well in different environments or different LiDAR sensors.

Instead of using dense semantic information estimated by neural networks~\cite{milioto2019iros, li2020arxiv}, a rather lighter solution has been proposed for long-term localization, which extracts only pole landmarks from point clouds.
There are usually two parts in pole-based approaches, pole extraction and pose estimation.
For pole extraction, Sefati \etal~\cite{sefati2017iv} first remove the ground plane from the point cloud and project the rest points on a horizontal grid. 
After that, they cluster the grid cells and fit a cylinder for each cluster. 
Finally, a particle filter with nearest-neighbor data association is used for pose estimation.
Weng \etal~\cite{weng2018rcar} and Schaefer \etal~\cite{schaefer2019ecmr} use similar particle filter-based methods to estimate the pose of the robot with different pole extractors. 
Weng \etal~\cite{weng2018rcar} discretize the space and extract poles based on the number of laser reflections in each voxel. 
Based on that, Schaefer \etal~\cite{schaefer2019ecmr} consider both the starting and end points of the scan and thus model the occupied and free space explicitly.
K{\"u}mmerle~\etal~\cite{kummerle2019icra} use a nonlinear least-squares optimization method to refine the pose estimation.

In contrast to the aforementioned approaches, we use a projection-based method and avoid the comparable costly processing of 3D point cloud data. Thus, our implementation is fast.
Besides, our approach uses range information directly without exploiting neural networks or deep learning, and thus, it generalizes well to different environments and different LiDAR sensors and does not require new training data when moving to different environments.

%%%%%%%%%%%%%%%%%%%%%%%%%%%%%%%%%%%%%%%%%%%%%%%%%%%%%%%%%%%%%%%%%%%%%%%%%%%%%%%%
\section{Our Approach}
\label{sec:main}

\begin{figure}
	\includegraphics[width=\linewidth]{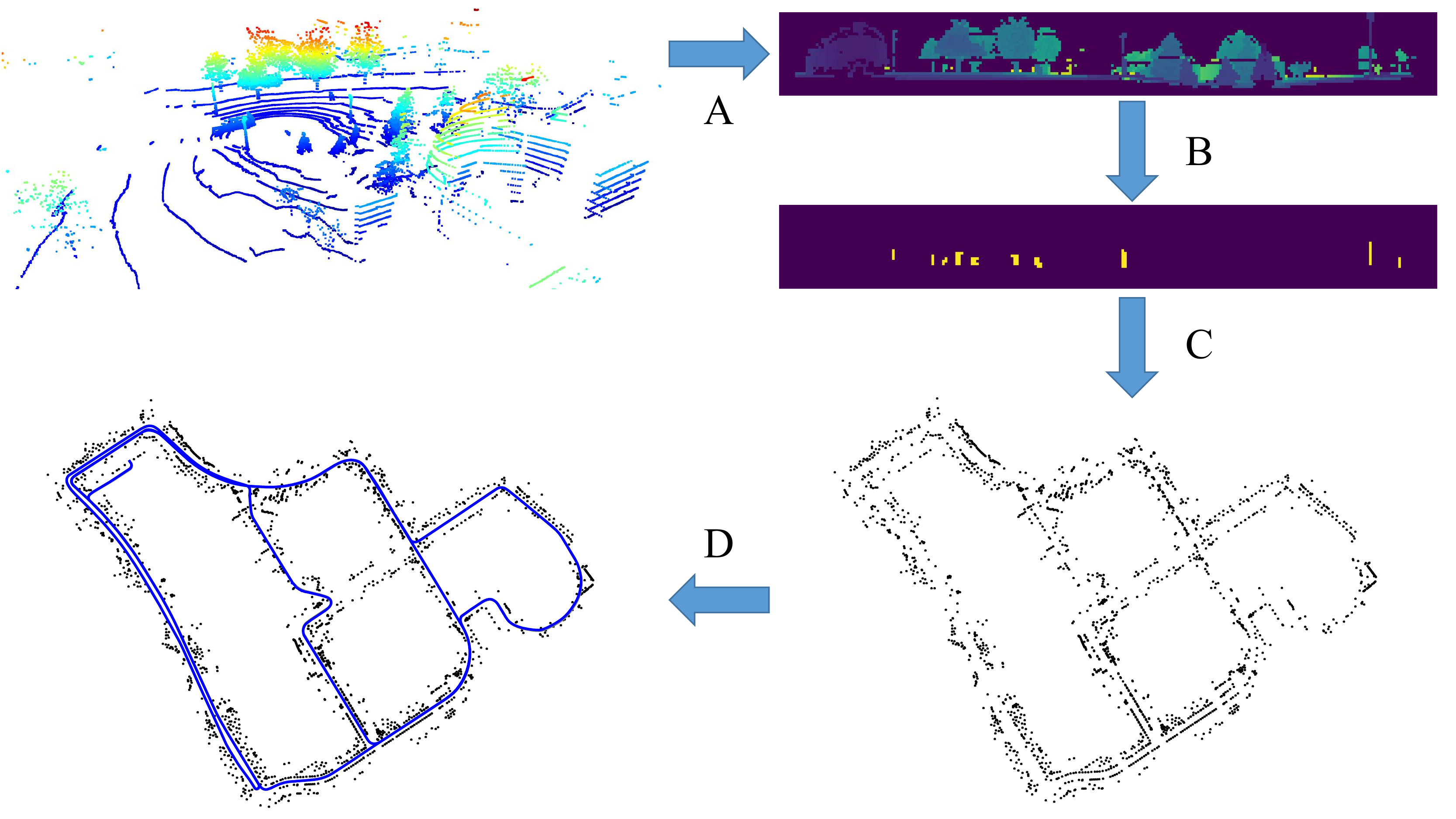}
	\caption{Overview of our approach. A. we project the LiDAR point cloud into a range image and B. extract poles in the image. C. based on the extracted poles, we then build a 2D global pole map of the environment. D. we finally propose a pole-based observation model for MCL to localize the robot in the map.}
	\label{fig:system}
  \end{figure}

In this paper, we propose a range image-based pole extractor for long-term localization using a 3D LiDAR sensor.
As shown in \figref{fig:system}, we first project the LiDAR point cloud into a range image (\secref{rangeGe}) and extract poles from it (\secref{poleEx}). 
Based on the proposed pole extractor, we then build a global pole map of the environment (\secref{map}).
In the localization phase, we extract poles online using the same extractor, and use a novel pole-based observation model for Monte Carlo localization (\secref{loc}).

\subsection{Range Image Generation\label{rangeGe}}
The key idea of our approach is to use range images generated from LiDAR scans for pole extraction. 
Following the prior work~\cite{chen2019iros, chen2021icra}, we utilize a spherical projection for range images generation. 
Each LiDAR point $\v{p}=(x, y, z)$ is mapped to spherical coordinates via a mapping $\Pi: \RR^3 \mapsto \RR^2$ 
and finally to image coordinates, as defined by
\begin{align}
	\left( \begin{array}{c} u \vspace{0.0em} \\ v \end{array}\right) & = \left(\begin{array}{cc} \frac{1}{2}\left[1-\arctan(y, x) \, \pi^{-1}\right]~\,~w   \vspace{0.5em} \\
			\left[1 - \left(\arcsin(z\, r^{-1}) + \mathrm{f}_{\mathrm{up}}\right) \mathrm{f}^{-1}\right] \, h\end{array} \right), \label{eq:projection}
\end{align}
where $(u,v)$ are image coordinates, $(h, w)$ are the height and width of the desired range image, $\mathrm{f}~{=}~\mathrm{f}_{\mathrm{up}}~{+}~\mathrm{f}_{\mathrm{down}}$ is the vertical field-of-view of the sensor, and $r~{=}~||\v{p}_i||_2$ is the range of each point.
This procedure results in a list of~$(u,v)$ tuples containing a pair of image coordinates for each $\v{p}_i$, which we use to generate our proxy representation. Using these indices, we extract for each $\v{p}_i$, its range~$r$, its $x$, $y$, and~$z$ coordinates, and store them in the image.

\subsection{Pole Extraction\label{poleEx}}
We extract poles based on the range images generated in the previous step. 
The general intuition behind our pole extraction algorithm is that the range values of the poles are usually significantly smaller than the backgrounds. 
Based on this idea and as specified in~\algref{alg:pole_extractor}, our first step is to cluster the pixels of the range image into different small regions based on their range values. 
We first pass through all pixels in the range image, from top to bottom, left to right. 

We put all pixels with valid range data in an open set~$\mathbf{O}$. To remove ground points, we ignore pixels under a certain height.  
For each valid pixel~$\mathbf{p}$, we check its neighbors including the left, right and below ones.
If there exists a neighbor with a valid value and the range differences between the current pixel and its neighbour is smaller than a threshold~$\epsilon$, we add the current pixel to a cluster set~$\mathbf{c}$ and remove it from the open set~$\mathbf{O}$. 
We do the same check iteratively with the neighbors until no neighbor pixel meets the above criteria, and we then get a cluster of pixels.
After checking all the pixels in~$\mathbf{O}$, we will get a set~$\mathbf{C}$ with several clusters and each cluster represents one object.
If the number of pixels in one cluster is smaller than a threshold $\zeta$, we regard it as an outlier and ignore it.

The next step is to extract poles from these objects using geometric constraints.
To this end, we exploit both the range information and the 3D coordinates~$(x, y, z)$ of each pixel.
We first check the aspect ratio of each cluster.
Since we are only interested in pole-like objects, whose height is usually larger than its width, we therefore throw away the cluster with aspect ratio~$h / w \textless 1$.
Another heuristic we use is the fact that a pole usually stands alone and has a significant distance from background objects.
$N_\mathit{SmallR}$ is the number of points in cluster~$\mathbf{c}$ whose range value is smaller than its neighbour outside~$\mathbf{c}$,
we throw away the cluster if $N_\mathit{SmallR}$ is smaller than $\delta$ times the number of all points in the cluster.

To exploit 3D coordinates~$(x, y, z)$ of each pixel, 
we calculate $\max(z) - \min(z)$ of each cluster and only take a cluster as a pole candidate if $\max(z) - \min(z) > \gamma$.
Besides, we are only interested in poles whose height is higher than~$\alpha$.
Based on experience, we also set a threshold~$\beta$ for the lowest position of the pole to filter outliers.
For each pole candidate, we then fit a circle using the~$x$ and~$y$ coordinates of all points in the cluster and get the center and the radius of that pole.
We filter out the candidates with too small or too large radiuses and candidates that connect to other objects by checking the free space around them.
After the above steps, we finally extract the positions and radiuses of poles. 

\begin{algorithm2e}[t]

    \small
    \SetAlgoLined
	\DontPrintSemicolon
	\KwInput{Range Image $\mathbf{I}_{range}$}
	\KwOutput{PoleParameters $\mathbf{P}$ with circle centers and radiuses}
	Let $\mathbf{O}$ be the set of all valid pixel coordinates in $\mathbf{I}_{range}$. \;
	\While{$\mathbf{O} \neq \emptyset$ }
	{	create a new $\mathbf{c}$ in $\mathbf{C}$;
	    $\mathbf{p} \gets \mathbf{O}[0]$; 
		$\mathbf{N} \gets neighbor(\mathbf{p})$ \;
		\While{$\mathbf{N} \neq \emptyset$}
		{
			\If{any $neighbor(\mathbf{p})  \in \mathbf{O}$ and $distance(neighbor(\mathbf{p}), \mathbf{p}) \textless \mathbf{\epsilon}$}
			{
				$\mathbf{c} \gets \mathbf{c} + \{\mathbf{p}\}$;
				$\mathbf{p} \gets \mathbf{N}[0]$\;
				$\mathbf{N} \gets \mathbf{N} + neighbor(\mathbf{p})$ \;				
			}
		}
		$N_\mathit{pixel} \gets $ the number of pixels in $\mathbf{c}$\;
		\If{$N_\mathit{pixel} \textless \zeta$}
		{
		    $\mathbf{C} \gets \mathbf{C} - \mathbf{c}$\;
		}
	}
	\ForEach{$\mathbf{c} \in \mathbf{C}$}
	{
		$w, h \gets width(\mathbf{c}), height(\mathbf{c})$\;
		$N_\mathit{SmallR} \gets $ the number of pixels in $\mathbf{c}$ whose range value is smaller than its neighbour outside $\mathbf{c}$ \;
		\If{$h / w \textless 1$ or $N_\mathit{SmallR} \textless \delta \cdot len(\mathbf{c}) $}
		{
			$\mathbf{C} \gets \mathbf{C} - \mathbf{c}$\;
		}
	}
	\ForEach{$\mathbf{c} \in \mathbf{C}$}
	{
		$\mathbf{x}, \mathbf{y}, \mathbf{z} \gets $ 3D coordinates of pixels in $\mathbf{c}$\;
		\If{$\max(\mathbf{z}) \textgreater \alpha$ and $\min(\mathbf{z}) \textless \beta$ and $(\max(\mathbf{z})-\min(\mathbf{z})) \textgreater \gamma$}
		{
			$x_\mathbf{c}, y_\mathbf{c}, r_\mathbf{c} \gets FitCircle(\mathbf{x},\mathbf{y})$ \;
			$N_\mathit{Free} \gets $ the number of the pixels in a small free space outside the radius of the pole\;
			\If{$r_\mathbf{c} \textless a$ and $r_\mathbf{c} \textgreater b$ and $N_\mathit{Free} \textless \mu \cdot len(\mathbf{z})$ } 
			{
				$\mathbf{P} \gets \mathbf{P} + \{ x_\mathbf{c}, y_\mathbf{c}, r_\mathbf{c} \}$
			}
		}
	}
  \caption{Pole Extraction Based on Range Image}
  \label{alg:pole_extractor}
\end{algorithm2e}

\subsection{Mapping\label{map}}
To build the 2D global pole map for localization, we following the same setup as introduced by  Schaefer \etal~\cite{schaefer2019ecmr}, splitting the ground-truth trajectory into shorter sections with equal length.
Since the provided poses are not very accurate for mapping~\cite{schaefer2019ecmr}, instead of aggregating a noisy submap, we only use the middle LiDAR scan of each section to extract poles.
We merge multiple overlapped pole detections by averaging over their centers and radiuses and apply a counting model to filter out the dynamic objects.
Only those poles that appear multiple times in continuous sections are regarded as real poles.

\subsection{Monte Carlo Localization\label{loc}}

Monte Carlo localization~(MCL) is commonly implemented using a particle filter~\cite{dellaert1999icra}. 
MCL realizes a recursive Bayesian filter estimating a probability density~$p(\v{x}_t\mid\v{z}_{1:t},\v{u}_{1:t})$ over the pose~$\v{x_t}$ given all observations~$\v{z}_{1:t}$ and motion controls~$\v{u}_{1:t}$ up to time~$t$. This posterior is updated as follows:
\begin{align}
  &p(\v{x}_t\mid\v{z}_{1:t},\v{u}_{1:t}) = \eta~p(\v{z}_t\mid\v{x}_{t}) \cdot\nonumber\\
  &\;\;\int{p(\v{x}_t\mid\v{u}_{t}, \v{x}_{t-1})~p(\v{x}_{t-1} \mid \v{z}_{1:t-1},\v{u}_{1:t-1})\ d\v{x}_{t-1}},
\label{eq:bayesian}
\end{align}
where~$\eta$ is a normalization constant, $p(\v{x}_t\mid\v{u}_{t}, \v{x}_{t-1})$~is the motion model, $p(\v{z}_t\mid\v{x}_{t})$~is the observation model, and $p(\v{x}_{t-1} \mid \v{z}_{1:t-1},\v{u}_{1:t-1})$~is the probability distribution for the prior state $\v{v}_{t-1}$.

In our case, each particle represents a hypothesis for the 2D pose $\v{x}_t = (x, y, \theta)_t$ of the robot at time~$t$. 
When the robot moves, the pose of each particle is updated based on a motion model with the control input~$\v{u}_t$ or the odometry measurements.
For the observation model, the weights of the particles are updated based on the difference between expected observations and actual observations.
The observations are the positions of the poles. We match the online observed poles with the poles in the map via nearest-neighbor search using a k-d tree. 
The likelihood of the $j$-th particle is then approximated using a Gaussian distribution:
\begin{align}
  p \left(\v{z}_{t} \mid \v{x}_{t} \right) &\propto \prod_{i}^{N} \exp{\left(-\frac{1}{2} \frac{{d\left(\v{z}^{i}_{t}, \v{z}^{i}_{j} \right)}^2}{\sigma^2_d}\right)},
\label{eq:sensor}
\end{align}
where~$d$ corresponds to the difference between the online observed pole~$\v{z}^{i}_{t}$ and matched pole in the map $\v{z}^{i}_{j}$ given the position of the  particle~$j$. $N$ is the number of matches of current scan.
We use the Euclidean distance between the pole positions to measure this difference. 
If the number of effective particles decreases below a threshold~\cite{grisetti2007tro}, the resampling process is triggered and particles are sampled based on their weights.

%%%%%%%%%%%%%%%%%%%%%%%%%%%%%%%%%%%%%%%%%%%%%%%%%%%%%%%%%%%%%%%%%%%%%%%%%%%%%%%%
\section{Experimental Evaluation}
\label{sec:exp}

The main focus of this work is an accurate and efficient pole extractor for long-term LiDAR localization.
We present our experiments to show the capabilities of our method and to
support our key claims, that our method is able to:
(i)~better extract poles compared to the baseline method,
(ii)~as a result, achieve better performance on long-term localization in different environments, and
(iii)~achieve online operation without using GPUs.

%%%%%%%%%%%%%%%%%%%%%%%%
\subsection{Pole Extractor Performance}

The first experiment evaluates the pole extraction performance of our approach and supports the claim that our range image-based method outperforms the baseline method in pole extraction. 

To the best of our knowledge, there is no specific public dataset available to evaluate pole extraction performance. 
To this end, we label the poles in session 2012-01-08 of NCLT dataset by hand and release this dataset for public research use. 
For the reason that the original NCLT ground-truth poses are inaccurate~\cite{schaefer2019ecmr}, the aggregated point cloud is a little blurry. Therefore, to create the ground-truth pole map of the environment, we partition the ground-truth trajectory into shorter segments of equal length. 
For each segment, we aggregate the point cloud together and use Open3D~\cite{zhou2018arXiv} to render and label the pole positions. 
We only label those poles with high certainty and ignore those blurry ones.
Besides our own labelled data, we also reorganize the SemanticKITTI~\cite{behley2019iccv} dataset sequence 00-10 by extracting the pole-like objects, \eg traffic signs, poles and trunk, and then clustering the point clouds to generate the ground-truth pole instances. 
We evaluate our method and compare it to the state-of-the-art pole-based LiDAR localization method proposed by Schaefer \etal~\cite{schaefer2019ecmr} in both datasets.

During the matching phase, we find the matches via nearest-neighbor search using a k-d tree with $1\,m$ distance bounds.
\tabref{tab:pole_extractor} summarizes the precision, recall and F1 score of our method and Schaefer \etal~\cite{schaefer2019ecmr} compared to the ground-truth pole map on both NCLT dataset and SemanticKITTI dataset. 
As can be seen, our method has better performance and extracts more poles in both environments.

\begin{table}[t]
\caption{Pole Extraction Accuracy on NCLT and KITTI datasets.}
\centering
\scalebox{1}{
\scriptsize{
\begin{tabular}{ll|ccc}
	\toprule
    Dataset & Method & Precision & Recall & F1 Score \\
	\midrule
	\multirow{2}{1cm}{NCLT} & Schaefer~\cite{schaefer2019ecmr} & 0.518  & 0.664 & 0.582 \\
	& Ours & $\mathbf{0.525}$  & $\mathbf{0.713}$ &  $\mathbf{0.605}$\\
	\midrule
	\multirow{2}{1cm}{Semantic KITTI}  & Schaefer~\cite{schaefer2019ecmr} &0.621 &0.380 & 0.455\\
	& Ours &$\mathbf{0.687}$ & $\mathbf{0.439}$&$\mathbf{0.515}$ \\
	\bottomrule
\end{tabular}
}
}
\label{tab:pole_extractor}
\end{table}

\subsection{Localization Performance}
The second experiment is presented to support the claim that our approach achieves higher accuracy on localization in different environments.
To assess the localization reliability and accuracy of our method, we use three datasets for evaluation, NCLT dataset~\cite{carlevaris-bianco2016ijrr}, MulRan dataset~\cite{kim2020icra}, and KITTI Odometry dataset~\cite{geiger2012cvpr}. Note that, these three datasets are collected in different environments (U.S., Korea, Germany) with different LiDAR sensors (Velodyne HDL-32E, Ouster OS1-64, Velodyne HDL-64E).
In the NCLT and MulRan dataset, the robot passes through the same place multiple times with month-level temporal gaps, hence ideal to test the long-term localization performance. We compare our methods to both pole-based method from Schaefer \etal~\cite{schaefer2019ecmr} and the range image-based method proposed by Chen \etal~\cite{chen2021icra}.
For the KITTI dataset, we follow the setup introduced by Schaefer \etal~\cite{schaefer2019ecmr} and compare our method with methods proposed by Schaefer \etal~\cite{schaefer2019ecmr}, K{\"u}mmerle~\etal~\cite{kummerle2019icra}, Weng~\etal~\cite{weng2018rcar} and Sefati~\etal~\cite{sefati2017iv}.
For all the experiments, we use the same setup as used in the baselines and report their results from the original work.

\subsubsection{Localization on the NCLT Dataset}

The NCLT dataset contains $27$ sessions with an average length of $5.5\,km$ and an average duration of $1.3\,h$ over the course of $15$ months.
The data is recorded at different times over a year, different weather and seasons, including both indoor and outdoor environments, and also lots of dynamic objects. 
The trajectories of different sessions have a large overlap. 
Therefore, it is an ideal dataset for testing long-term localization in urban environments. 

We first build the map following the setup introduced by Schaefer \etal~\cite{schaefer2019ecmr}, which uses the laser scans and the ground-truth poses of the first session. Since in later sessions, the robot sometimes moves into unseen places for the first session, we therefore also use those scans whose position is $10\,m$ away from all previously visited poses to build the map. 
During localization, we use $1000$ particles and use the same initialization as~\cite{schaefer2019ecmr} by uniformly sampling positions around the first ground-truth pose within a $2.5\,m$ circle. The orientations are uniformly sampled from $-5$ to $5$ degree. 
We resample particles if the number of effective particles is less than $0.5$. To get the pose estimation, we use the average poses of the best $10\,\%$ of the particles.

\begin{figure}
	\includegraphics[width=0.95\linewidth]{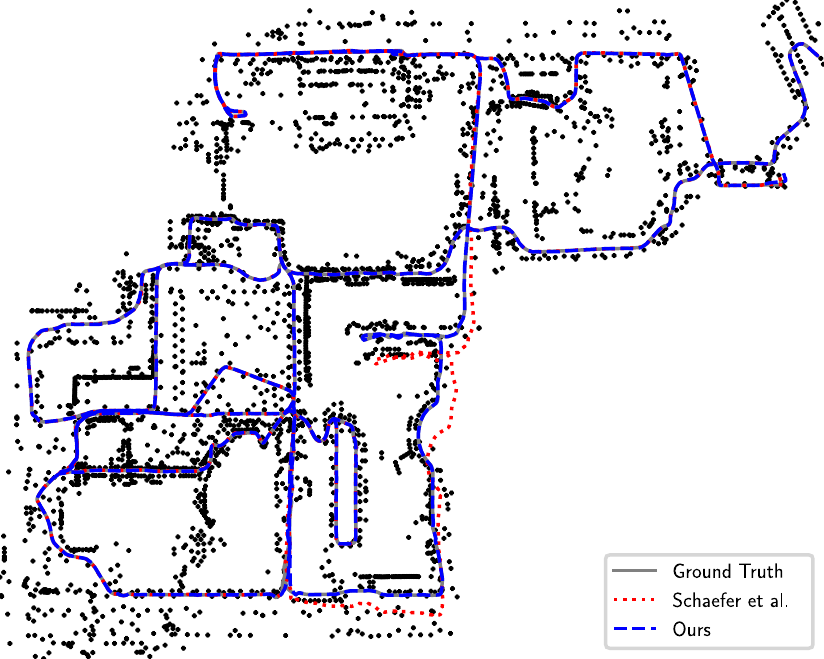}
	\caption{Comparison of localization results of Schaefer \etal~\cite{schaefer2019ecmr} and our method in session 2012-02-23 on NCLT dataset.
	The black dots are the poles on the map. The grey line is the ground-truth trajectory. The blue line is our result and the red one is of the baseline method.
	As can be seen, Schaefer's method loses the track of the robot in some places, while our method always tracks the correct robot poses with respect to the ground truth.}
	\label{fig:loc_compare}
  \end{figure}

\tabref{tab:nclt_results} shows the position and orientation
errors for every session. We run the localization $10$ times and compute the average means and RMSEs to the ground-truth trajectory. The results show that our method surpasses Schaefer \etal~\cite{schaefer2019ecmr} in almost all sessions with an average error of $0.173\,m$. 
Besides, in session 2012-02-23, the baseline method fails to localize resulting in an error of $2.470\,m$, while our method never loses track of the robot position (\figref{fig:loc_compare}). 
This is because our pole extractor can robustly extract poles even in an environment where there are fewer poles.

\begin{table}[t]
  \caption{Localization Results on MulRan Dataset.}
  \centering
  \small
  \scalebox{0.9}{
  \begin{tabular}{C{3cm}ccc}
    \toprule
     & Schaefer~\cite{schaefer2019ecmr} & Chen~\cite{chen2021icra}  & ours  \\
	 \midrule
	 $\RMSE_{\mathrm{pos}}$~[\si{m}]  & 1.82 &0.83 & $\mathbf{0.48}$ \\
	 $\RMSE_{\mathrm{ang}}$~[\si{\degree}]  & 0.56 &3.14 & $\mathbf{0.27}$ \\
	\bottomrule
  \end{tabular}
}
  \label{tab:mulran_results}
\end{table}

\begin{table*}
	\centering
	\small
	\scalebox{0.9}{
	\setlength{\tabcolsep}{3pt}
	\begin{tabular}{l
	        S[round-mode=places, round-precision=1]
			S[round-mode=places, round-precision=3]
			S[round-mode=places, round-precision=3]
			S[round-mode=places, round-precision=3]
			S[round-mode=places, round-precision=3]
			S[round-mode=places, round-precision=3]
			S[round-mode=places, round-precision=3]
			S[round-mode=places, round-precision=3]
			S[round-mode=places, round-precision=3]}
		\toprule
		Session Date 
			& $f_{\mathrm{map}}$
			& \multicolumn{2}{c}{$\Delta_{\mathrm{pos}}$}
			& \multicolumn{2}{c}{$\RMSE_{\mathrm{pos}}$}
			& \multicolumn{2}{c}{$\Delta_{\mathrm{ang}}$}
			& \multicolumn{2}{c}{$\RMSE_{\mathrm{ang}}$} \\
			& [\si{\%}]
			& \multicolumn{2}{c}{[\si{m}]}
			& \multicolumn{2}{c}{[\si{m}]}
			& \multicolumn{2}{c}{[\si{\degree}]}
			& \multicolumn{2}{c}{[\si{\degree}]} \\
		\toprule
			& 
			& \text{Schaefer~\cite{schaefer2019ecmr}} & Ours
			& \text{Schaefer~\cite{schaefer2019ecmr}} & Ours
			& \text{Schaefer~\cite{schaefer2019ecmr}} & Ours
			& \text{Schaefer~\cite{schaefer2019ecmr}} & Ours \\
		\midrule
		2012-01-08 & 100.0000000000 & $\mathrm{0.130}$ &$\mathbf{0.115}$ & $\mathrm{0.178}$ &$\mathbf{0.146}$ & $\mathrm{0.663}$ & $\mathbf{0.626}$ & $\mathrm{0.857}$ & $\mathbf{0.810}$\\
		2012-01-15 & 8.46853977048  & $\mathrm{0.156}$ & $\mathbf{0.146}$ & $\mathrm{0.225}$ &  $\mathbf{0.204}$ & $\mathrm{0.760}$ & $\mathbf{0.750}$ & $\mathrm{0.999}$ & $\mathbf{0.982}$\\
		2012-01-22 & 5.09090909091  & $\mathrm{0.172}$ &   $\mathbf{0.149}$ & $\mathrm{0.222}$ &  $\mathbf{0.190}$ & $\mathrm{0.939}$ &  $\mathbf{0.911}$ & $\mathrm{1.291}$ & $\mathbf{1.238}$\\
		2012-02-02 & 0.355956336023  & $\mathrm{0.155}$ &  $\mathbf{0.136}$& $\mathrm{0.205}$ &$\mathbf{0.172}$ & $\mathrm{0.720}$ & $\mathbf{0.699}$& $\mathrm{0.975}$ & $\mathbf{0.922}$ \\
		2012-02-04 & 0.106298166357  & $\mathrm{0.144}$ & $\mathbf{0.134}$  & $\mathrm{0.195}$ &$\mathbf{0.167}$ & $\mathrm{0.684}$ & $\mathbf{0.671}$& $\mathrm{0.903}$ & $\mathbf{0.876}$\\
		2012-02-05 & 0.54114994363  & $\mathrm{0.148}$ & $\mathbf{0.138}$ & $\mathrm{0.210}$ & $\mathbf{0.206}$ & $\mathbf{0.690}$ & $\mathrm{0.694}$ & $\mathrm{0.947}$ & $\mathbf{0.937}$ \\
		2012-02-12 & 0.822833633325  & $\mathrm{0.269}$ & $\mathbf{0.253}$ & $\mathrm{1.005}$ &  $\mathbf{1.002}$ & $\mathrm{0.802}$ &  $\mathbf{0.786}$ & $\mathrm{1.040}$ &  $\mathbf{1.019}$\\
		2012-02-18 & 0.815738963532  & $\mathrm{0.149}$ & $\mathbf{0.131}$  & $\mathrm{0.221}$ &$\mathbf{0.175}$ & $\mathrm{0.699}$ & $\mathbf{0.676}$& $\mathrm{0.938}$ & $\mathbf{0.905}$ \\
		2012-02-19 & 0.048111618956  & $\mathrm{0.148}$ &$\mathbf{0.137}$ & $\mathrm{0.194}$ & $\mathbf{0.183}$& $\mathrm{0.704}$ & $\mathbf{0.692}$  & $\mathrm{0.944}$ & $\mathbf{0.923}$ \\
		2012-03-17 & 0.000000000000  & $\mathrm{0.149}$ &$\mathbf{0.137}$ & $\mathrm{0.191}$ &$\mathbf{0.174}$ & $\mathrm{0.830}$ &  $\mathbf{0.798}$& $\mathrm{1.062}$ &$\mathbf{1.031}$ \\
		2012-03-25 & 0.000000000000  & $\mathrm{0.200}$ &  $\mathbf{0.178}$ & $\mathrm{0.262}$ & $\mathbf{0.235}$& $\mathrm{1.418}$ &  $\mathbf{1.379}$& $\mathrm{1.836}$ &  $\mathbf{1.789}$ \\
		2012-03-31 & 0.000000000000  & $\mathrm{0.143}$ &$\mathbf{0.135}$ & $\mathrm{0.184}$ &$\mathbf{0.176}$ & $\mathrm{0.746}$ & $\mathbf{0.729}$ & $\mathrm{0.973}$ &$\mathbf{0.936}$ \\
		2012-04-29 & 0.000000000000  & $\mathrm{0.170}$ &$\mathbf{0.154}$ & $\mathrm{0.251}$ & $\mathbf{0.222}$   & $\mathrm{0.829}$ & $\mathbf{0.820}$ & $\mathrm{1.079}$ &$\mathbf{1.069}$ \\
		2012-05-11 & 5.46702623192  & $\mathrm{0.161}$ & $\mathbf{0.132}$& $\mathrm{0.225}$ &$\mathbf{0.163}$ & $\mathrm{0.773}$ &  $\mathbf{0.747}$ & $\mathrm{0.998}$ & $\mathbf{0.965}$\\
		2012-05-26 & 0.378340033105  & $\mathrm{0.158}$ & $\mathbf{0.142}$ & $\mathrm{0.217}$ &   $\mathbf{0.183}$& $\mathrm{0.690}$ &  $\mathbf{0.672}$ & $\mathrm{0.889}$ & $\mathbf{0.871}$\\
		2012-06-15 & 0.403669724771  & $\mathrm{0.180}$ & $\mathbf{0.145}$& $\mathrm{0.238}$ & $\mathbf{0.186}$& $\mathrm{0.659}$ & $\mathbf{0.646}$ & $\mathrm{0.879}$ &  $\mathbf{0.874}$\\
		2012-08-04 & 0.273000273  & $\mathrm{0.210}$ & $\mathbf{0.169}$& $\mathrm{0.340}$ &$\mathbf{0.230}$ & $\mathrm{0.884}$ &  $\mathbf{0.843}$  & $\mathrm{1.143}$ & $\mathbf{1.093}$\\
		2012-08-20 & 3.81451009723  & $\mathrm{0.189}$ & $\mathbf{0.156}$& $\mathrm{0.264}$ & $\mathbf{0.207}$  & $\mathrm{0.711}$ &  $\mathbf{0.688}$  & $\mathrm{0.941}$ & $\mathbf{0.906}$\\
		2012-09-28 & 0.34965034965  & $\mathrm{0.206}$ &  $\mathbf{0.171}$ & $\mathrm{0.311}$ &  $\mathbf{0.241}$& $\mathrm{0.731}$ &  $\mathbf{0.726}$ & $\mathrm{0.952}$ & $\mathbf{0.949}$\\
		2012-10-28 & 1.42405063291 & $\mathrm{0.217}$ &$\mathbf{0.185}$ & $\mathrm{0.338}$ &$\mathbf{0.281}$ & $\mathrm{0.693}$ &$\mathbf{0.680}$   & $\mathrm{0.919}$ & $\mathbf{0.909}$\\
		2012-11-04 & 2.53441802253  & $\mathrm{0.257}$ & $\mathbf{0.208}$& $\mathrm{0.456}$ &$\mathbf{0.317}$ & $\mathrm{0.746}$ & $\mathbf{0.718}$  & $\mathrm{0.996}$ & $\mathbf{0.973}$\\
		2012-11-16 & 2.72922416437  & $\mathrm{0.403}$ &  $\mathbf{0.296}$& $\mathrm{0.722}$ & $\mathbf{0.435}$& $\mathrm{1.467}$ & $\mathbf{1.403}$& $\mathrm{2.031}$ &  $\mathbf{1.919}$ \\
		2012-11-17 & 0.364678301641  & $\mathrm{0.243}$ &$\mathbf{0.201}$ & $\mathrm{0.377}$ &  $\mathbf{0.323}$& $\mathrm{0.686}$ & $\mathbf{0.685}$ & $\mathrm{0.959}$ & $\mathbf{0.948}$\\
		2012-12-01 & 0.00000000000 & $\mathrm{0.266}$ &  $\mathbf{0.226}$  & $\mathrm{0.492}$ &$\mathbf{0.429}$ & $\mathrm{0.674}$ &$\mathbf{0.665}$  & $\mathrm{0.930}$ & $\mathbf{0.887}$ \\
		2013-01-10 & 0.00000000000 & $\mathrm{0.217}$ & $\mathbf{0.187}$ & $\mathrm{0.278}$ &$\mathbf{0.226}$ & $\mathrm{0.689}$ & $\mathbf{0.627}$& $\mathrm{0.911}$ &$\mathbf{0.806}$ \\
		2013-02-23 & 0.00000000000 & $\mathrm{2.470}$  & $\mathbf{0.236}$ & $\mathrm{5.480}$ & $\mathbf{0.567}$ & $\mathrm{1.083}$ & $\mathbf{0.592}$ & $\mathrm{1.769}$ & $\mathbf{0.846}$\\
		2013-04-05 & 0.00000000000 & $\mathrm{0.365}$ &$\mathbf{0.295}$ & $\mathrm{0.920}$ &$\mathbf{0.869}$ & $\mathrm{0.654}$ & $\mathbf{0.642}$& $\mathbf{1.028}$ &$\mathrm{1.036}$ \\
\toprule
		Average	& & $\mathrm{0.284}$& $\mathbf{0.174}$& $\mathrm{0.526}$&$\mathbf{0.293}$ &$\mathrm{0.801}$ & $\mathbf{0.761}$& $\mathrm{1.081}$&$\mathbf{1.016}$\\
		\bottomrule
	\end{tabular}
}
	\caption{
		Results of our experiments with the NCLT dataset compared to Schaefer~\cite{schaefer2019ecmr}, averaged over ten localization runs per session.
		The variables $\Delta_{\textrm{pos}}$ and $\Delta_{\textrm{ang}}$ denote the mean absolute errors in position and heading, respectively, $\RMSE_{\textrm{pos}}$ and $\RMSE_{\textrm{ang}}$ represent the corresponding root mean squared errors, while $f_{\textrm{map}}$ denotes the fraction of lidar scans per session used to build the  reference map.}
	\label{tab:nclt_results}
\end{table*}

\subsubsection{Localization on the MulRan Dataset}
In MulRan dataset, we use KAIST 02 sequence (collected on 2019-08-23) to build the global map and use KAIST 01 sequence (collected on 2019-06-20) for localization.
\tabref{tab:mulran_results} shows the location and yaw angle RMSE
errors on MulRan Dataset. 
As can be seen, our method consistently achieves a better performance than both baseline methods~\cite{schaefer2019ecmr,chen2021icra}.

\subsubsection{Localization on the KITTI Dataset}
For the KITTI dataset, we follow the setup introduced in Schaefer \etal~\cite{schaefer2019ecmr}, where sequence number 9 data is used for both mapping and localization.
\tabref{tab:kitti_results} shows the localization results.
Our method consistently achieves a better performance than all four baseline methods.

\begin{table*}
	\centering
	\small
	\scalebox{0.9}{
	\begin{tabular}{lccccccccc}
		\toprule
		Approach
			& $\Delta_{\textrm{pos}}$
			& $\RMSE_{\textrm{pos}}$
			& $\Delta_{\textrm{lat}}$
			& $\sigma_{\textrm{lat}}$
			& $\Delta_{\textrm{lon}}$
			& $\sigma_{\textrm{lon}}$
			& $\Delta_{\textrm{ang}}$
			& $\sigma_{\textrm{ang}}$
			& $\RMSE_{\textrm{ang}}$ \\
			& [\si{m}]
			& [\si{m}]
			& [\si{m}]
			& [\si{m}]
			& [\si{m}]
			& [\si{m}]
			& [\si{\degree}]
			& [\si{\degree}]
			& [\si{\degree}] \\
		\midrule
		K{\"u}mmerle et~al.~\cite{kummerle2019icra}
			& \tablenum[table-format=1.3]{0.12}
			& ---
			& \tablenum[table-format=1.3]{0.07}
			& ---
			& \tablenum[table-format=1.3]{0.08}
			& ---
			& \tablenum[table-format=1.3]{0.33}
			& ---
			& --- \\
		Weng et~al.~\cite{weng2018rcar}
			& ---
			& ---
			& ---
			& \tablenum[table-format=1.3]{0.082}
			& ---
			& \tablenum[table-format=1.3]{0.164}
			& ---
			& \tablenum[table-format=1.3]{0.329}
			& --- \\
		Sefati et~al.~\cite{sefati2017iv}
			& ---
			& \tablenum[table-format=1.3]{0.24}
			& ---
			& ---
			& ---
			& ---
			& ---
			& ---
			& \tablenum[table-format=1.3]{0.68} \\		
		Schaefer et~al.~\cite{schaefer2019ecmr}
			& \tablenum[table-format=1.3]{0.096}
			& \tablenum[table-format=1.3]{0.111}
			& $\mathbf{0.061}$
			& \tablenum[table-format=1.3]{0.075}
			& \tablenum[table-format=1.3]{0.060}
			& \tablenum[table-format=1.3]{0.067}
			& \tablenum[table-format=1.3]{0.133}
			& \tablenum[table-format=1.3]{0.188}
			& \tablenum[table-format=1.3]{0.214} \\
		Ours
			& $\mathbf{0.091}$
			& $\mathbf{0.106}$
			& \tablenum[table-format=1.3]{0.066}
			& $\mathbf{0.067}$
			& $\mathbf{0.046}$
			& $\mathbf{0.058}$
			& $\mathbf{0.084}$
			& $\mathbf{0.102}$
			& $\mathbf{0.102}$ \\
		\bottomrule
	\end{tabular}
	}
	\caption{
		Comparison of the accuracies of Sefati \etal. and Schaefer \etal.'s method and the proposed localization approach on the KITTI dataset.
		The results of Weng \etal. and K{\"u}mmerle \etal. are not directly comparable and are stated for qualitative analysis only.}
	\label{tab:kitti_results}
\end{table*}

%%%%%%%%%%%%%%%%%%%%%%%%
\subsection{Runtime}

This experiment has been conducted to support the claim that our
approach runs online at the sensor frame rate without using GPUs. 
We compare our method to the baseline method proposed by Schaefer \etal~\cite{schaefer2019ecmr}. As reported in their paper, the baseline method takes an average of $1.33\,s$ for pole extraction on a PC using \emph{a dedicated GPU}.
We tested our method without using a GPU and our method only needs $0.09\,s$ for pole extraction and all MCL steps take less than $0.1\,s$ yielding a run time faster than the LiDAR frame rate of $10\,Hz$.

%%%%%%%%%%%%%%%%%%%%%%%%%%%%%%%%%%%%%%%%%%%%%%%%%%%%%%%%%%%%%%%%%%%%%%%%%%%%%%%%
\section{Conclusion}
\label{sec:conclusion}

In this paper, we presented a novel range image-based pole extraction approach for online long-term LiDAR localization.
Our method exploits range images generated from LiDAR scans.
This allows our method to process point cloud data rapidly and run online.
We implemented and evaluated our approach on different datasets
and provided comparisons to other existing techniques and supported
all claims made in this paper. The experiments suggest that our method can extract
more poles in the environments accurately and achieve better performance in 
long-term localization tasks. Moreover, we release our implementation and pole dataset for other researchers to evaluate their algorithms. 
In the future, we plan to explore the usage of other features such as road markings, curb and intersection features, to improve the robustness of our method.

%%%%%%%%%%%%%%%%%%%%%%%%%%%%%%%%%%%%%%%%%%%%%%%%%%%%%%%%%%%%%%%%%%%%%%%%%%%%%%%%
%% Future work: Use only if applicable -- but if so, use the following
%% sentence to start:
% Despite these encouraging results, there is further space for improvements. 

%%%%%%%%%%%%%%%%%%%%%%%%%%%%%%%%%%%%%%%%%%%%%%%%%%%%%%%%%%%%%%%%%%%%%%%%%%%%%%%%
% Only if applicable
%\section*{Acknowledgments}
%We thank XXX for fruitful discussions and for \dots

\bibliographystyle{plain_abbrv}

% All new citations should go to new.bib. The file glorified.bib should go
% be the one from the ipb server. After paper or related work has been
% written merge the entries from new.bib to glorified.bib ON THE SERVER,
% replace the glorified.bib in this repository and empty the new.bib
\bibliography{glorified,new}

\end{document}